%% file: main.tex
\documentclass[sigconf,nonacm]{acmart}

\usepackage[utf8]{inputenc}
\usepackage{amsmath}
\usepackage{multirow}
\usepackage{microtype}
\usepackage{boldline}
\usepackage{hyperref}
\usepackage{xcolor}
\usepackage{diagbox}
\usepackage{makecell}
\usepackage{wasysym}
\usepackage{enumitem}

\begin{document}

\title{Measuring Graph-to-Graph Semantic Similarity in Knowledge Graphs: An Empirical Evaluation of Knowledge Graph Embeddings}

\author{Seungryeol Baek}
\affiliation{
  \institution{Department of Artificial Intelligence, Sungkyunkwan University}
  \city{Suwon}
  \country{Republic of Korea}
}

\author{Wooseok Sim}
\affiliation{
  \institution{Department of Artificial Intelligence, Sungkyunkwan University}
  \city{Suwon}
  \country{Republic of Korea}
}

\author{Hogun Park}
\affiliation{
  \institution{Department of Artificial Intelligence, Sungkyunkwan University}
  \city{Suwon}
  \country{Republic of Korea}
}

\input{abstract}


\maketitle

\input{intro}
\input{related_work}

\input{preliminaries}
\input{methods}
\input{experiment}
\input{results}
\input{discussion}
\input{conclusion}

\begin{acks}
This work was supported by the Institute of Information \& Communications Technology Planning \& Evaluation (IITP) and the National Research Foundation of Korea (NRF), both funded by the Ministry of Science and ICT (MSIT), under Grant Nos. RS-2025-24803185, RS-2019-II190421, and IITP-2025-RS-2020-II201821.
\end{acks}

\bibliographystyle{ACM-Reference-Format}
\bibliography{bib}

\appendix
\input{appendix}

\end{document}

%% file: abstract.tex
\begin{abstract}
A Knowledge Graph (KG) represents facts as structured triples and is widely used to organize relational knowledge across diverse domains.
Just as textual information ranges from words and sentences to complete documents, KG information can be interpreted at multiple levels, from entities, relations, and triples to subgraphs and entire KGs.
However, existing KG embedding methods mainly focus on entities, relations, and triples, leaving graph-level semantics largely unaddressed.
Conventional graph-level methods, which typically compare graphs based on structural patterns, are also insufficient because structural similarity alone cannot guarantee semantic similarity between KGs.
To evaluate how well different methods capture such graph-level semantic information, we study graph-to-graph semantic similarity, which determines whether a pair of KGs represents semantically corresponding underlying information.
To obtain reliable ground-truth correspondences, we construct a semantic matching dataset by modifying text documents, extracting KGs from both original and modified documents, and transferring their known correspondences to KG pairs.
We compare text-based, structure-based, and KG embedding-based approaches on each dataset.
For the KG embedding-based approach, we introduce two scoring functions: \textit{EmbPairSim}, which uses maximal pairwise entity similarity, and \textit{AvgEmbSim}, which uses a frequency-weighted centroid.
Experiments on WikiText-2 and CC-News show that \textit{EmbPairSim} achieves up to 5.3 pp higher MRR than Sentence-BERT while using substantially fewer parameters.
These results suggest that KGE representations can serve as compact and effective signals for graph-to-graph semantic similarity in KGs.
Our code is available at \url{https://github.com/SeungRyeolBaek/KG-to-KG-Semantic-Similarity}.
\end{abstract}

%% file: intro.tex
\section{Introduction}\label{sec:intro}

A \emph{Knowledge Graph} (KG) represents a fact as a triple $(h,r,t)$ that connects a \textbf{h}ead entity $h$ to a \textbf{t}ail entity $t$ via a typed \textbf{r}elation $r$—e.g.,~\texttt{(\textit{Paris}, located\_in, \textit{France})}.
Due to their ability to organize complex relational knowledge in a structured form, KGs have become a widely adopted knowledge representation across diverse domains.
Examples include encyclopedic knowledge in Wikidata~\cite{wikidata}, biomedical knowledge in BioKG~\cite{BioKG}, legal knowledge in LegalKG~\cite{LegalKG}, scholarly knowledge in ORKG~\cite{ScholarlyKG}, event-centric knowledge in EventKG~\cite{EventKG}, and visual relationships in scene graphs~\cite{VisualGenome}.

Information stored in KGs can be analyzed at multiple levels, ranging from entities and relations to triples, subgraphs, and entire KGs.
Entities and relations denote individual units and the types of connections between them.
Triples represent specific relational facts.
Subgraphs, formed by multiple triples, can capture broader semantic units, such as events, research contributions, or scene regions.
An entire KG can represent a complete information source, such as encyclopedic, biomedical, scholarly, visual, or document-derived knowledge.
Appendix~\ref{sec:multi_level_kg_examples} illustrates this multi-level view across representative KG domains.
This view is analogous to textual information, which can be studied at different levels ranging from individual words and sentences to complete documents.

Although semantic representations of KGs have been extensively studied, most existing approaches focus on local components such as entities, relations, and triples rather than subgraphs or entire KGs.
Triple-centered language-model approaches such as KG-BERT~\cite{yao2019kgbert}, KEPLER~\cite{wang2021kepler}, and RLKB~\cite{FAN201731} verbalize individual triples and encode them with Transformers~\cite{NIPS2017_3f5ee243}, but rely primarily on triple-level semantics rather than the graph structure of an entire KG.
Knowledge Graph Embedding (KGE) models, including transductive approaches~\cite{TransE,DistMult,ComplEx,RotatE} and inductive approaches~\cite{ingram}, incorporate both semantic and structural information.
Nevertheless, these methods mainly model local KG components, such as entities, relations, and triples, rather than the semantic information represented by an entire KG.

Addressing KGs at the graph level may appear to align with conventional graph-level tasks.
However, conventional graph-level tasks usually characterize graphs based on structural patterns, such as topology, connectivity, and substructures.
This structural perspective is useful, but it is not sufficient for KGs, because KGs are constructed by organizing semantic information.
In particular, structural similarity alone cannot guarantee that two KGs represent similar information.

\input{framework/framework}
Together, these limitations indicate that graph-level semantic information in KGs remains insufficiently addressed.
Therefore, we aim to evaluate how well different types of methods can capture the semantic information represented by an entire KG.
A natural evaluation setting for this purpose is graph-to-graph semantic similarity, where semantically corresponding KGs should be identified among other candidate graphs.
Specifically, we formulate the task as determining whether two KGs represent the same underlying information at the graph level, rather than whether they merely share similar structures.
Figure~\ref{fig:framework} exemplifies the task.  
Given a query KG, $G_q$ (top), we rank candidate graphs $\{G_i\}$ such that the one encoding the \emph{same real‑world situation} surfaces first.  
An effective measure requires semantic awareness to bridge lexical variation, structural sensitivity to account for graph-level organization, and scalability to handle document-sized KGs.

To evaluate this task, however, we need reliable ground-truth correspondences between KGs.
Such correspondences are difficult to obtain from existing KGs, which are often already processed into graph form, making it difficult to trace their original sources or determine whether two independently constructed KGs should be regarded as semantically similar.
To obtain controllable semantic correspondences, we instead start from text documents.
By applying meaning-preserving modifications, such as lexical substitution and paraphrasing, each original document can be paired with a modified document that expresses the same underlying information.
After converting both the original and modified documents into KGs, the known document-level correspondence can be transferred to the resulting KG pair.
This allows us to construct evaluation data in which each query KG has a clear ground-truth counterpart.

Based on this idea, we construct a \textbf{semantic matching} dataset:
(i) documents from \textit{WikiText‑2} and \textit{CC‑News} are parsed into KGs with an LLM pipeline; (ii) each document is paraphrased at six lexical/structural intensities; (iii) KGs are re‑extracted from the paraphrases; and (iv) every query graph must retrieve its true counterpart among hundreds of distractors.  
We report Hits@5, MRR, and NDCG across the graph pairs.

Using each dataset, we empirically compare text-based, structure-based, and KG embedding-based approaches for graph-to-graph semantic similarity.
For the KG embedding-based approach, we introduce two complementary,  scoring functions that lift off-the-shelf KGEs to the graph level:
(1) \textbf{EmbPairSim}: computes maximal pairwise cosine similarity between entities (and optionally relations) in two graphs, preserving fine-grained correspondences;
(2) \textbf{AvgEmbSim}: forms a single frequency-weighted centroid vector per graph, enabling more efficient retrieval.

For comparison, we use Sentence-BERT (SBERT)~\cite{sentencebert} as the text-based approach and Graph Kernel methods~\cite{base_kernel,wl-kernel} as the structure-based approach.
For SBERT, KG triples are verbalized before being used as input.
Across our dataset, \textit{EmbPairSim} achieves up to 5.3 pp higher MRR than Sentence-BERT while using substantially fewer parameters.
These results suggest that KGE representations can serve as compact and effective signals for graph-to-graph semantic similarity in KGs.

Our key contributions are as follows:
\begin{itemize}[leftmargin=*]
  \item \textbf{Task \& dataset. To the best of our knowledge, we release the first benchmark specifically targeting document-derived KG-to-KG semantic retrieval under controlled paraphrasing conditions.}
  \item \textbf{Lightweight KGE aggregation.  We propose \textit{EmbPairSim} and \textit{AvgEmbSim}, two scoring functions that require \emph{no} retraining and run in sub‑second time.}
  \item \textbf{Empirical findings. \textit{EmbPairSim} achieves up to 5.3 pp higher MRR than Sentence-BERT while using an order of magnitude fewer parameters; an ablation study shows that frequency weighting generally helps, mean-centering is important for INGRAM-based similarity, and relation embeddings can introduce noise.}
\end{itemize}

%% file: framework/framework.tex
\begin{figure}[t]
  \centering
  \includegraphics[width=1\linewidth]{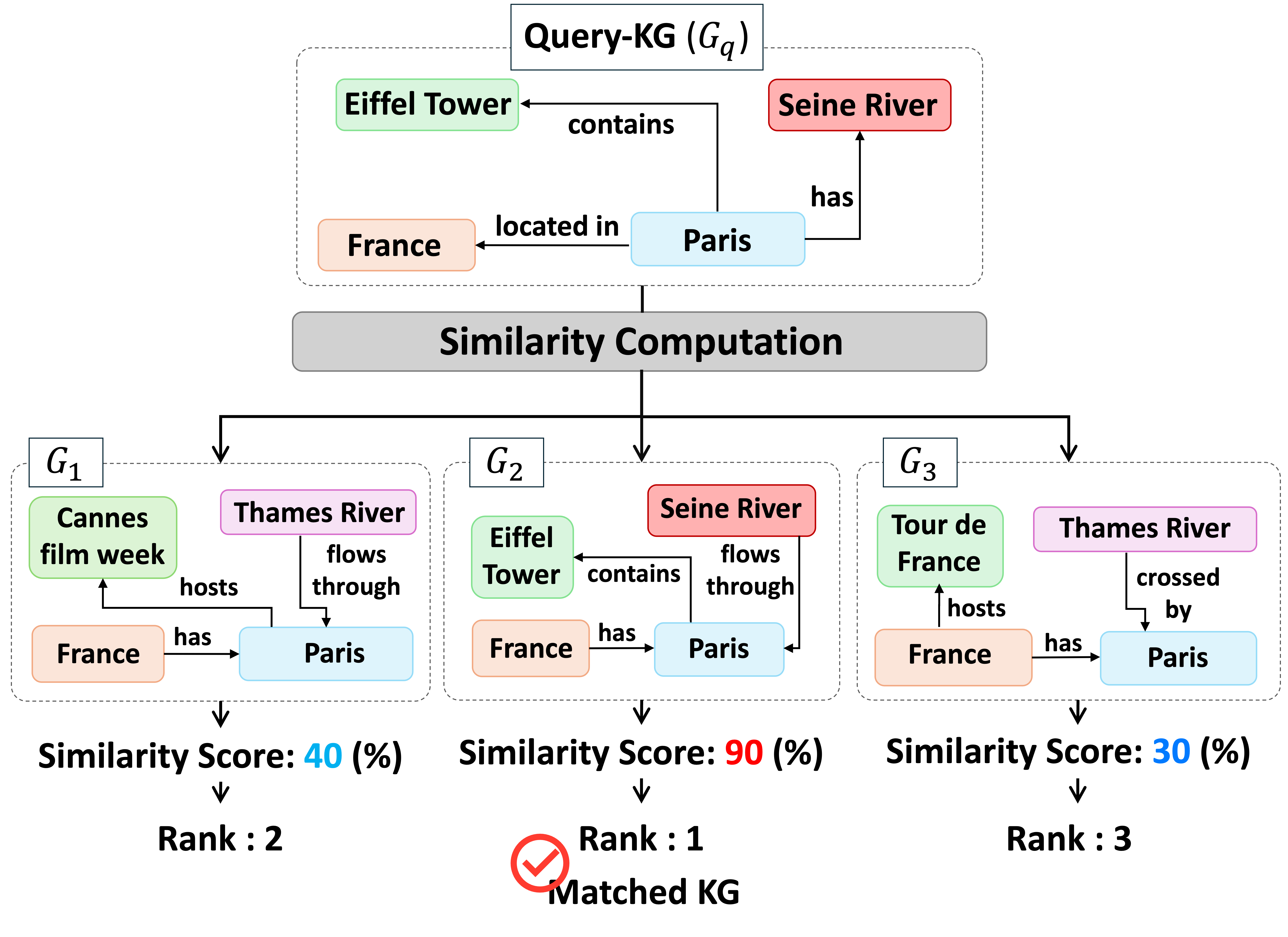}
  \vspace{-9mm}
  \caption{Overview of our \textbf{semantic matching} task.}
  \label{fig:framework}
  \vspace{-3mm}
\end{figure}

%% file: related_work.tex
\section{Related Work}\label{sec:related_work}
\subsection{Language-based Representations} \label{sec:semantic_embedding}
Word embeddings like Word2Vec \cite{word2vec} represent words with similar contexts close together in a continuous vector space. 
% , enabling fine-grained semantic similarity.
Sentence-level models such as Sentence-BERT (SBERT)~\cite{sentencebert} extend this to entire sentences or documents by capturing contextual information across tokens. These embeddings support semantic similarity, retrieval, and clustering.
In our work, we leverage SBERT to measure semantic similarity between verbalized Knowledge Graphs (KGs).
% , treating them as natural language documents. 
This approach serves as a language-based baseline for assessing graph-level similarity.

\subsection{Structure‑based Approaches} \label{graph kernels}
Graph-based methods compare graph structures by identifying matching substructures or edit operations. Early approaches such as Graph Edit Distance (GED)~\cite{graph-edit-distance} and Ullmann's subgraph matching~\cite{subgraph-matching} provide exact structural similarity but are computationally expensive and do not scale well.
Graph kernels address this by mapping graphs to high-dimensional feature spaces based on statistical patterns, enabling efficient similarity computation. Examples include the Vertex and Edge Histogram Kernels~\cite{base_kernel}, which use label distributions, and the Weisfeiler-Lehman (WL) kernel~\cite{wl-kernel}, which iteratively relabels nodes to capture multi-level structural information. These kernels balance structural detail and scalability, making them practical for large graphs.
In our work, we use graph kernels as structure-based baselines to evaluate how well surface-level structural features approximate semantic similarity between KGs.

\subsection{Knowledge Graph Embedding (KGE) Methods}\label{sec:kge}
Knowledge Graph Embedding (KGE) methods map entities and relations into continuous vector spaces while preserving structural properties. They support tasks such as link prediction, entity classification, and graph completion.
TransE~\cite{TransE} models relations as translations in embedding space, while DistMult~\cite{DistMult} uses a multiplicative scoring function based on element-wise interactions. ComplEx~\cite{ComplEx} extends this by using complex-valued embeddings for richer relational patterns, and RotatE~\cite{RotatE} models relations as rotations in complex space. These models are transductive and require all entities and relations during training.
In contrast, INGRAM \cite{ingram} follows an inductive paradigm: a graph neural network derives embeddings for previously unseen entities and relations directly from their local graph structure. It relies exclusively on structural context and does not retain any entity or relation vectors produced during training.
% It dynamically infers embeddings at test time with a DistMult-like scoring mechanism.
These KGE models are analogous to word embeddings but additionally encode graph-specific structure. We use them as structure-aware baselines to assess graph-level semantic similarity.

%% file: preliminaries.tex
\section{Preliminaries}\label{preliminaries}
\subsection{Notations}\label{notation}
A Knowledge Graph (KG) is defined as a tuple $G = (\mathcal E, \mathcal R, \mathcal T)$, where $\mathcal E$ is the set of entities ($|\mathcal E|$ denotes its size), $\mathcal R$ is the set of relations ($|\mathcal R|$ denotes its size), and $\mathcal T \subseteq \{(h, r, t)~|~h, t \in \mathcal E, r \in \mathcal R\}$ is the set of triples, and $|\mathcal T|$ is their total count. Each triple $(h, r, t)$ represents a directed relation from head entity $h$ to tail entity $t$ via relation $r$. Thus, a KG can be viewed as a directed labeled graph with entities as nodes and triples as edges.

\subsection{Knowledge graph as a Document}\label{knowledge_graph}
Textual data exhibits a hierarchical structure: words form sentences, sentences combine into paragraphs, and paragraphs compose a document.  
Although a knowledge graph arranges entities and relations in a non-linear network without a fixed reading order, we can view it through the same hierarchy: entities and relations act as words, each triplet as a sentence, and subgraphs—or the entire graph—as paragraphs or even a full document. Viewing KGs this way allows us to apply document-level operations, such as similarity measurement, while capturing both structure and semantics.

\subsection{Graph Kernel-based  Approach}\label{sec:kernel} % graph set 도입 필요 없음, graph 2개만, KG 정의 도입 필요
As a baseline for measuring similarity between KGs, we use graph kernel methods. Given two KGs, $G$ and $G'$, their similarity is computed as:
\begin{equation}
    S_\text{kernel} = K(G,G^\prime).
\end{equation}
This approach relies purely on structural features. We use two kernels in our main experiments: the Vertex Histogram Kernel~\cite{base_kernel}, and the Weisfeiler-Lehman (WL) kernel~\cite{wl-kernel}.
The Vertex Histogram Kernel compares distributions of discrete entity identifiers without using learned semantic embeddings. 
The WL Kernel further incorporates local graph topology by iteratively updating node labels, providing richer structural similarity than simple histograms.

\section{Similarity Evaluation Setup}\label{sec:preprocess}
\paragraph{Problem formulation.}
To objectively evaluate the graph-to-graph semantic similarity task, we formulate a semantic matching problem between two KG sets.
Let $\mathcal G = \{G_1, \ldots, G_m\}$ be a set of candidate KGs and let $\mathcal G' = \{G'_1, \ldots, G'_n\}$ be a set of query KGs, where $m \ge n$.
This formulation assumes that each $G'_i$ represents the same underlying information as $G_i$, while it is distinct from other candidate graphs $G_j$ ($j \neq i$).
Based on this assumption, the goal is to identify, for each query KG $G'_i \in \mathcal G'$, the candidate KG $G_i \in \mathcal G$ that semantically matches it.

\paragraph{Construction of aligned KG pairs.}
To obtain controllable semantic correspondences, we first construct semantically corresponding document pairs and then convert each document into a KG.
Specifically, given a document set $\mathcal D = \{D_1, \ldots, D_m\}$, we first select a subset $\mathcal D_n = \{D_1, \ldots, D_n\} \subset \mathcal D$ ($m \ge n$).
We then generate modified documents $\mathcal D' = \{D'_1, \ldots, D'_n\}$ using a document modification function that preserves the meaning of the original document.
Each modified document $D'_i$ preserves the underlying information of its original document $D_i$ while remaining distinct from other documents $D_j$ ($j \neq i$).

We then convert both original and modified documents into KGs using the same LLM pipeline.
Specifically, using a same Large Language Model (LLM)~\cite{openai2023gpt35} pipeline, we extract a KG $ G_i$ from each original document $D_i$  and a KG $G'_i$ from each modified document $D'_i$, forming $\mathcal G = \{G_1, \ldots, G_m\}$ and $\mathcal G' = \{G'_1, \ldots, G'_n\}$, respectively
Since $G_i$ and $G'_i$ are extracted from semantically corresponding documents, the pair $(G'_i, G_i)$ provides a ground-truth semantic match.
As a result, we obtain KG sets with controllable semantic correspondences that are suitable for the semantic matching problem

\paragraph{Evaluation protocol.}
Given the constructed KG sets $\mathcal G$ and $\mathcal G'$, we compute pairwise similarity scores between each query KG $G'_i \in \mathcal G'$ and every candidate KG $G_j \in \mathcal G$.
These scores form a similarity matrix whose $(i,j)$-th entry indicates the predicted semantic similarity between $G'_i$ and $G_j$.
For each query KG $G'_i$, candidate KGs are ranked by their similarity scores.
The ranking is evaluated by checking whether the ground-truth counterpart $G_i$ is ranked highly among all candidates.
We report standard retrieval metrics, including Hits, NDCG, and MRR.

%% file: methods.tex
\section{Methods}\label{sec:method}
\subsection{Embedding-Based Pairwise Similarity}\label{sec:embpairsim}
We propose \textbf{Embedding-based Pairwise Similarity (\textit{EmbPairSim})} to measure semantic similarity between two Knowledge Graphs (KGs), $G=(\mathcal E, \mathcal R, \mathcal T)$ and $G^\prime=(\mathcal E^\prime, \mathcal R^\prime, \mathcal T^\prime)$.
Instead of exact label matching, which is often too rigid, we use embeddings-vector representations of entities and relations-to compare elements based on their semantic meaning.
%, similar to word embeddings in Natural Language Processing (NLP)~\cite{word2vec}.

\subsubsection{Embedding Generation}
We generate embeddings for all entities and relations using an embedding function $\text{Embed}(\cdot)$. For entities, this yields:
\begin{align}
    \textbf{e}_i=\text{Embed}(e_i), ~\text{for}~ e_i\in\mathcal E, \\
    \textbf{e}_j^\prime=\text{Embed}(e_j^\prime), ~\text{for}~ e_j^\prime\in\mathcal E^\prime.
\end{align}
Each $\textbf{e}_i$ and $\textbf{e}_j'$ encodes the semantic meaning of the corresponding entity in $G$ and $G'$, respectively.
The embedding function $\text{Embed}(\cdot)$ can be instantiated either by a transductive KGE model, where entity embeddings for both $G$ and $G'$ are learned jointly in a shared embedding space by training on the union of their triples, or by an inductive KGE model that derives embeddings from graph structure using pretrained parameters.

% Similarly, for relations, we obtain:
% \begin{align}
%     \textbf{r}_i=\text{Embed}(r_i), ~\text{for}~ r_i\in\mathcal R, \\
%     \textbf{r}_j^\prime=\text{Embed}(r_j^\prime), ~\text{for}~ r_j^\prime\in\mathcal R^\prime.
% \end{align}
% Each embedding $\textbf{r}_i$ represents the meaning of the relation $r_i$ in the knowledge graph $G$ and each embedding $\textbf{r}'_j$ represents the meaning of the relation $r'_j$ in the knowledge graph $G'$. 

\subsubsection{Mean Centering}
To align embeddings from both graphs, we apply \textbf{mean centering} by computing their joint mean vector and subtracting it from each entity and relation embedding before similarity calculation.
\begin{equation}
\begin{aligned}
    \mu_E &= \frac{1}{|\mathcal E| + |\mathcal E'|}\!
             \left( \sum_{i=1}^{|\mathcal E|} \textbf{e}_i
                  + \sum_{j=1}^{|\mathcal E'|} \textbf{e}'_{\,j} \right),\\
    \textbf{e}_i &\leftarrow \textbf{e}_i - \mu_E,\quad
    \textbf{e}'_{\,j} \leftarrow \textbf{e}'_{\,j} - \mu_E.
\end{aligned}
\end{equation}
We use a joint mean vector from both graphs to preserve the relative positions between embeddings across the two KGs while spreading them around the origin, thereby increasing the contrast of cosine similarity distributions during comparison.

\subsubsection{Stacking Embedding Vectors to Matrix}
We stack these embedding vectors to obtain matrices:
\begin{equation}
    \begin{aligned}
        \textbf{E} = \left[ \textbf{e}_1, \textbf{e}_2, ... , \textbf{e}_{|\mathcal{E}|}  \right], \\
        \textbf{E}' = \left[ \textbf{e}'_1, \textbf{e}'_2, ... , \textbf{e}'_{|\mathcal{E}'|}  \right], \\
    \end{aligned}
\end{equation}
where $\mathbf{E} \in \mathbb{R}^{d \times |\mathcal{E}|}$ and $\mathbf{E}' \in \mathbb{R}^{d \times |\mathcal{E}'|}$ contain the entity embeddings of $G$ and $G'$, respectively. These matrices are used in the subsequent pairwise similarity computation.

\subsubsection{Pairwise Similarity Calculation}
We compute pairwise \textbf{cosine similarities} by first applying column-wise normalization to embedding matrices $\mathbf{E}$ and $\mathbf{E}'$ and then taking the dot product between the normalized embedding matrices.
\begin{equation}\label{eq_pairSim}
    \mathbf{Sim}^{\mathbf E}
    =
    \hat{\mathbf E}^{\top}\hat{\mathbf E}',
    \quad
    \hat{\mathbf E}
    =
    \mathbf E \operatorname{diag}(\mathbf E^{\top}\mathbf E)^{-\frac{1}{2}},
    \quad
    \hat{\mathbf E}'
    =
    \mathbf E' \operatorname{diag}({\mathbf E'}^{\top}\mathbf E')^{-\frac{1}{2}},
\end{equation}
where $\operatorname{diag}(\cdot)$ extracts the diagonal elements of a square matrix and forms a diagonal matrix from them.
Each value in the matrix $\mathbf{Sim}^\mathbf{E}$ is the cosine similarity between an entity from $G$ and one from $G'$.

\subsubsection{Handling Size Differences}
% To fairly compare KGs of different sizes, we base similarity on the smaller graph.
When comparing KGs of different sizes, we measure how well the entities in the smaller KG are covered by entities in the larger KG.
If similarity is instead aggregated from the larger KG to the smaller KG, multiple entities in the larger KG are inevitably matched to the same entity in the smaller KG, which can make additional information in the larger KG negatively affect the similarity score.
For each entity in the smaller KG, we find its most similar entity in the larger KG by taking the maximum similarity value from \(\mathbf{Sim}^\mathbf{E}\).
Each row of \(\mathbf{Sim}^\mathbf{E}\) corresponds to an entity in $G$ and each column to one in $G'$.
Selecting the maximum per row or column yields the closest match for entities in one graph to the other.
Multiple entities can still share the same nearest matching entity.
\begin{equation}
S_V =
\begin{cases}
  \{\max_{i\in \mathcal E},~\mathbf{Sim}^\mathbf{E}_{i,j}| j\in \mathcal E^\prime\}, &\text{if } |\mathcal E| > |\mathcal E^\prime|, \\
  \{\max_{j\in \mathcal E^\prime},\mathbf{Sim}^\mathbf{E}_{i,j}| i\in \mathcal E\}. &\text{otherwise}.
\end{cases}
\end{equation}
Here, $\mathbf{Sim}^\mathbf{E}_{i,j}$ denotes the similarity between the $i$-th entity in $G$ and the $j$-th entity in $G'$.
$S_V$ collects the highest similarity scores for entities in the smaller KG to their best matches in the larger KG.

\subsubsection{Computing the Final Similarity Score}
Finally, the overall similarity between $G$ and $G'$ is defined as the fraction of scores $s$ in $S_V$ exceeding a threshold $t$:
\begin{equation}\label{eq_sPair}
    S_{\text{pair}} =  { \frac{|\{s\in S_{V}| s>t\}|}
     {{|S_{V}|}} }.
\end{equation}
A higher $S_{\text{pair}}$ indicates greater semantic similarity between two KGs.

\subsection{Averaged Embedding-Based Similarity}\label{sec:avgembsim}
EmbPairSim preserves the full information in KG embeddings but must compute and aggregate pair-wise similarities for all entity pairs, which becomes costly as the graph grows. Moreover, it does not produce a single representation of a KG.
To obtain a single graph-level representation and avoid this quadratic overhead, we propose \textbf{Averaged Embedding-based Similarity (\textit{AvgEmbSim})}, which represents each graph by the frequency-weighted average of its embeddings; this summary vector provides a candidate proxy for the whole graph whose efficiency we will explicitly evaluate, and comparing weighted versus plain averages lets us test whether entity frequency matters in KGs as term frequency does in text.

% EmbPairSim computes similarity without discarding information in KG embeddings but does not provide a graph-level representation. To obtain such a representation, we introduce \textbf{Averaged Embedding-based Similarity (\textit{AvgEmbSim})}, which uses the frequency-weighted average of entity and relation embeddings as a single vector.
% Because averaging can omit detail, we evaluate its performance to identify this loss compared to EmbPairSim. 
% We also compare the frequency-weighted and simple averages in our experiments to assess whether entity frequency affects importance, as word frequency does in text.
% We propose \textbf{Averaged Embedding-based Similarity (\textit{AvgEmbSim})}, which represents a KG as a single vector by taking a frequency-weighted average of its entity and relation embeddings. 
% This captures the overall semantic meaning while reflecting the importance of each element.

\subsubsection{Frequency-Weighted Averaging}
% To reflect the importance of each entity and relation
To reflect the importance of each entity, we weight their embeddings by frequency, the number of triples in the knowledge graph that contain each entity, similar to word frequency in NLP. The frequency-weighted average embeddings are:
\begin{equation}\label{eq:avg_sim}
\begin{aligned}
    \bar {\textbf{e}} = { \frac{\sum_{e\in \mathcal{E}}\text{freq}(e)~\text{Embed}(e)}
    {\sum_{e\in \mathcal{E}}\text{freq}(e)} },
    \bar {\textbf{e}^\prime} = { \frac{\sum_{e^\prime\in \mathcal{E}^\prime}\text{freq}(e^\prime)~\text{Embed}(e^\prime)}
     {\sum_{e^\prime\in \mathcal{E}^\prime}\text{freq}(e^\prime)}},
\end{aligned}
\end{equation}
where $\text{freq}(e)$ denotes the frequency of entity $e$.

\subsubsection{Similarity Calculation}
Finally, we compute the cosine similarity between the averaged entity embeddings of $G$ and $G'$:
\begin{equation}\label{eq_sAvg}
    S_{\text{avg}} = { \frac{\bar {\textbf{e}}\cdot \bar {\textbf{e}^\prime} } {||\bar {\textbf{e}} ||~|| \bar {\textbf{e}^\prime}||}}.
\end{equation}

%% file: experiment.tex
\section{Experiments}
\subsection{Experimental Settings}
\subsubsection{Datasets \& KG Extraction}\label{sec:dataset}
We use two datasets: \textit{WikiText-2}~\cite{wikitext-2} (645 Wikipedia documents) and \textit{CC-News}~\cite{mackenzie2020ccnews} (550 news articles). 
% Together, they form the corpus $\mathcal D$.
We treat each dataset as a separate document collection and construct $\mathcal D$, $\mathcal D^\prime$, $\mathcal G$, and $\mathcal G^\prime$ independently from each dataset. 
Knowledge Graphs (KGs) are extracted using the LLMGraphTransformer from LangChain~\cite{chase2022langchain} with GPT-3.5-turbo~\cite{openai2023gpt35}, yielding graph sets $\mathcal G$ and $\mathcal G'$. 
We use the default prompt and extraction pipeline provided by the LLMGraphTransformer without additional prompt engineering or task-specific modifications.
All steps are performed independently for \textit{WikiText-2} and \textit{CC-News}.

\subsubsection{Document Modification} \label{sec:paraphrasing}
(1) \textbf{Synonym Replacement}: words are randomly selected and replaced with synonyms from WordNet~\cite{wordnet}. (2) \textbf{Context Replacement}: BERT~\cite{bert} is used to select contextually important words, which are then replaced with synonyms from WordNet~\cite{wordnet}. 
(3) \textbf{DIPPER Paraphraser}: generates diverse paraphrases with the DIPPER model \cite{dipper_paraphraser}. We categorize the modified document sets by method: \textbf{Synonym}, \textbf{Context}, and \textbf{DIPPER}. For \textbf{Synonym} and \textbf{Context}, modification strengths \textbf{30} (\%) and \textbf{60} (\%) indicate the fraction of tokens changed. For \textbf{DIPPER}, we test \textbf{60/0} and \textbf{60/20}, 
where the format $L/O$ denotes lexical diversity ($L$) and order diversity ($O$). These variations test the impact of different paraphrasing strategies on KG extraction and similarity. For each dataset, we randomly sample 200 documents and generate modified versions to form $\mathcal{D}'$. 

\subsubsection{Evaluation Metrics}\label{sec:metric}
We assess similarity performance using Hits@5, Mean Reciprocal Rank (MRR), and Normalized Discounted Cumulative Gain (NDCG). Hits@5 measures how often the correct KG appears in the top 5, MRR captures the average rank of the correct match, and NDCG evaluates ranking quality considering both relevance and position. Together, these metrics indicate how well the scores reflect true semantic similarity.

\subsubsection{Baselines}\label{sec:baseline}
We use text-based, graph kernel-based, and KG embedding-based baselines to compare how well each approach captures semantic similarity. For the embedding-based baselines, including Word2Vec, FastText, and transductive KGE models, we mainly followed the standard/default hyperparameter settings provided in the original papers or official implementations.
As a result, the transductive KGE models were evaluated under the same hyperparameter settings, including embedding dimension, margin, batch size, and learning rate.

\begin{itemize}[leftmargin=*]
    \item \textbf{Text-based Similarity.}  
    Each KG is verbalized into OpenIE sentences (e.g., $\langle h,r,t\rangle\!\to\!$ “$h$ $r$ $t$.”) with NLTK~\cite{bird2009nltk}, and graph similarity is computed from sentence embeddings given by the pretrained Sentence-BERT (SBERT) checkpoint \texttt{all-mpnet-base-v2}~\cite{allmpnet}.
    Moreover, in Table~\ref{tab:we_pair} we adopted two word embedding models, Word2Vec\cite{word2vec} and FastText\cite{mikolov2018fasttext}.
    For a non-pretrained setting, we train both models on the OpenIE sentences for 10 epochs, using 200-dimensional vectors, a window size of $=5$, negative samples $=5$, and min\_count$=2$ with uniform initialization for both models.
    In the pretrained configuration, we employ the 300-dimensional Google News Word2Vec embeddings \cite{mikolov2013word2vec}, and the FastText embeddings trained on the Wikipedia 2017, UMBC WebBase, and News Crawl corpora \cite{mikolov2018fasttext}.

    \item \textbf{Graph Kernel-based Similarity.} 
    We use the Vertex Histogram (VH) and Weisfeiler-Lehman (WL) kernels~\cite{base_kernel, wl-kernel} implemented with GraKeL~\cite{grakel} to compare the graphs based on node and edge label distributions.

    \item \textbf{Transductive KG Embedding-based Similarity.} 
    We used TransE~\cite{TransE}, DistMult~\cite{DistMult}, ComplEx~\cite{ComplEx}, and RotatE~\cite{RotatE} trained on the combined graphs of all the compared graphs. For ComplEx and RotatE, we use the real-valued entity embedding vectors returned by the PyG implementation when computing cosine similarity. We use an embedding dimension of 32, a margin of 2, a batch size of 128, and a learning rate of 0.01 with early stopping. We used initial embeddings in their original papers—uniform initialization for TransE and RotatE, and Xavier-uniform initialization for DistMult and ComplEx. The threshold for \textit{EmbPairSim} is determined via grid search and set to $t=0.8$.
    
    \item \textbf{Inductive KG Embedding-based Similarity.} 
    We use INGRAM \cite{ingram} as an inductive KGE model.
    We adopt the pretrained model trained on NELL-995 provided by the official code of~\cite{ingram} with an embedding dimension of 32. The threshold for \textit{EmbPairSim} is selected by grid search and set to $t=0.95$.
    Unlike transductive models, INGRAM derives context solely from the structural patterns and does not reference individual entities encountered during training. Consequently, its performance indicates how much contextual information can be captured using structural patterns alone.
\end{itemize}

%{0.70,0.75,0.8,0.85,0.9,0.95\}

%% file: results.tex
\input{results/main_result}
\subsection{Results}\label{sec:results}
\subsubsection{Overall Performance Comparisons}
The comparative results for the models outlined in Section~\ref{sec:baseline} are presented in Table~\ref{tab:combined_results}.
As detailed in Section~\ref{sec:model_comparison}, RotatE achieves the best performance among all transductive embedding models in \textit{EmbPairSim} and \textit{AvgEmbSim}. 
The \textbf{Emb size} column confirms that low-dimensional KG embeddings are more parameter-efficient than language-model ones. On the other hand, \textit{EmbPairSim}’s pairwise entity matching scales poorly on entity-dense sets like WikiText, whereas \textit{AvgEmbSim} uses a single graph embedding and is therefore much more efficient than SBERT or \textit{EmbPairSim}.

\paragraph{The Role of Semantic Information in Enhancing Similarity Detection}
\textit{EmbPairSim} outperforms both graph kernels, highlighting the advantage of semantic embeddings over purely statistical structural methods. This suggests that incorporating semantic information significantly improves graph similarity detection.

\paragraph{Comparative Performance of \textit{EmbPairSim} and SBERT}
\textit{EmbPairSim} achieves the highest overall performance, even surpassing SBERT in most cases. This shows that KG embeddings capture semantic meaning more effectively than text-based embeddings. However, under the \textbf{DIPPER} paraphrasing setting, SBERT performs better, indicating a limitation of \textit{\textit{EmbPairSim}} when handling substantial lexical and syntactic changes.

\paragraph{Effectiveness of Structural Information in Inductive Models}
\textit{EmbPairSim} and \textit{AvgEmbSim} perform similarly on both RotatE and INGRAM. Despite INGRAM's fully inductive setting, its performance is comparable to transductive models, suggesting that structural information is sufficient for capturing semantic meaning in KGs.
However, although INGRAM achieves almost the same \textit{EmbPairSim} performance as RotatE, it requires a substantially higher similarity threshold (0.95 versus 0.80 for RotatE), suggesting that INGRAM produces entity embeddings with generally higher cosine similarities and less dispersion in the embedding space than RotatE. 

\paragraph{Performance Gap between Vertex Histogram and Weisfeiler-Lehman Kernels}
The Vertex Histogram (VH) kernel slightly outperforms the Weisfeiler-Lehman (WL) Kernel. This appears to stem from WL's relabeling stage, which merges entity and relation labels, thereby blurring their distinction. Supporting this, we observed that WL using either the Vertex or Edge Histogram as its base yields identical results. Similarly, \textit{EmbPairSim}--based solely on entity embeddings--outperforms models that mix entity and relation information, highlighting the importance of preserving entity-level signals in semantic similarity tasks.

\paragraph{Limitations under Structural Paraphrasing}
While \textit{EmbPairSim} and \textit{AvgEmbSim} generally outperform text-based baselines, they show clear limitations when faced with substantial lexical and structural variations. The most notable performance drop is observed with the \textbf{DIPPER} Paraphraser, which introduces both large-scale vocabulary changes and global sentence reordering. Unlike the more localized word substitutions in the \textbf{Synonym} and \textbf{Context} settings, these changes disrupt surface patterns while preserving underlying semantics.
This degradation stems from the fact that both \textit{EmbPairSim} and \textit{AvgEmbSim} represent KGs through local statistics without capturing the graph’s relational structure or topology at a higher level. 
Although KGE-based methods consistently outperform SBERT in overall accuracy, their reliance on simple aggregation limits their ability to detect semantic alignment when meaning is preserved but surface forms are rearranged.
Modeling higher-order relational structure and graph-level semantics may therefore provide a more robust way to handle substantial paraphrastic transformations.
As a result, they remain vulnerable to paraphrastic variations that obscure local signals while maintaining global intent.

\input{analysis/kge_model/figure} 
\subsubsection{Selecting KGE Models for EmbPairSim and AvgEmbSim.}
\label{sec:model_comparison}
We evaluate four transductive KG embedding models—TransE, DistMult, ComplEx, and RotatE—as back-ends for \textit{EmbPairSim} and \textit{AvgEmbSim}, and summarize their NDCG performance across different paraphrasing settings in Figure~\ref{fig:models}. RotatE consistently delivers the highest scores, plausibly because its representation of relations as rotations in complex space aligns with the cosine similarity intrinsic to both metrics, retaining angular semantics. TransE performs worst; its purely translational assumption fails to provide the fine-grained relational separation required for reliable similarity estimation. DistMult and ComplEx behave similarly under \textit{EmbPairSim} due to their shared bilinear form, yet ComplEx slightly outperforms DistMult in \textit{AvgEmbSim}, reflecting its capacity to model asymmetric relations. Overall, models with multiplicative or angular inductive biases (RotatE, ComplEx) produce more effective embeddings for both pairwise and averaged similarity calculations.

\input{results/ablation_study/AvgEmbSim}
\input{results/ablation_study/EmbPairSim-INGRAM}
\subsubsection{Ablation Study}
\paragraph{Frequency of Elements Influences the Semantic Representation of a Knowledge Graph}
As part of the ablation study, we weight each embedding by its occurrence frequency instead of computing a simple average. Table~\ref{tab:ablation_avg} shows that this frequency-weighted approach generally improves performance. These findings suggest that the frequency of entities plays a crucial role in capturing the overall semantic information of a KG, similar to how word frequency aids in understanding text documents.

\paragraph{Mean-Centering Enhances Entity-Level Similarity}
Overall, mean-centering has little impact when \textit{EmbPairSim} employs transductive KGE models.
However, Table~\ref{tab:ablation_pair_ingram} shows that without mean-centering, the INGRAM-based \textit{EmbPairSim} fails to distinguish graphs.
This suggests that INGRAM produces entity vectors that are compressed into a narrow subspace.
This behavior may arise because INGRAM is pretrained on a single large KG.
Such pretraining captures structural patterns in the source graph, but when applied to the much smaller KGs in our experiments, it can map many entities into a narrow region of the embedding space.
Mean-centering alleviates this collapse by removing the global bias in the embedding space, making relative differences between entity vectors more visible for cosine-based matching.

\input{results/relation/EmbPairSim}
\subsubsection{Relation Embedding Considerations.}
Extending from prior experiments based solely on entity embeddings, we observe that including relation embeddings degrades similarity performance rather than improving it. For \textit{EmbPairSim}, we compute pairwise cosine similarities between relation embeddings in the same way as entities (Eq.~\eqref{eq_pairSim}), and include the result in the final score (Eq.~\eqref{eq_sPair}). For \textit{AvgEmbSim}, we calculate the frequency-weighted mean vector for relations (Eq.~\eqref{eq:avg_sim}) and add its cosine similarity to that of the entity vectors (Eq.~\eqref{eq_sAvg}). As shown in Table~\ref{tab:relation_pair_avg}, incorporating relation embeddings consistently degraded performance. This suggests that relation vectors may introduce noise rather than helpful signals, likely because semantic information from relations is already implicitly captured during entity embedding training.

\input{results/LM/EmbPairSim/trained}
\subsubsection{Comparing KG Embeddings with Word Embeddings.}
Because KG embeddings act as distributional representations, we can likewise apply \textit{EmbPairSim} by representing entities with word embeddings. 
Table~\ref{tab:we_pair} contrasts \textit{EmbPairSim} when entities are represented by KG embeddings versus word embeddings.
The results show that KGE-based embeddings are more effective to capture KG-to-KG semantic similarity than word embeddings.
This suggests that explicitly encoding entities through relational structure is more suitable for comparing KGs than relying on word embeddings learned from sequential textual contexts.
A likely reason is that KGE directly exploits graph-structured relations among entities, whereas word embeddings capture relational signals indirectly from sequential co-occurrence.

\begin{table}[t]
\centering
\small
\caption{Runtime comparison of graph-to-graph similarity methods.}
\label{tab:runtime_comparison}
\begin{tabular}{lr}
\toprule
Method & Runtime (s) \\
\midrule
VH Kernel & 1.569 \\
WL Kernel & 3.770 \\
SBERT & 12.374 \\
EmbPairSim (RotatE) & 0.381 \\
AvgEmbSim (RotatE) & 0.228 \\
\bottomrule
\end{tabular}
\end{table}

\subsubsection{Runtime Efficiency.}
Table~\ref{tab:runtime_comparison} reports the runtime of each graph-to-graph similarity method on WikiText-2 under the synonym replacement setting.
Among the baselines, VH Kernel and WL Kernel require 1.569 and 3.770 seconds, respectively, while SBERT requires 12.374 seconds due to sentence-level encoding over verbalized KG triples.
In contrast, the KGE-based scoring functions show substantially lower runtime.
EmbPairSim with RotatE completes the similarity computation in 0.381 seconds, and AvgEmbSim with RotatE further reduces the runtime to 0.228 seconds.
This result supports the efficiency of the proposed KGE-based scoring functions, especially AvgEmbSim, which represents each KG with a single frequency-weighted centroid instead of computing pairwise entity similarities.
Together with the performance results in Table~\ref{tab:combined_results}, these runtime results indicate that KGE-based graph-to-graph similarity can provide an efficient alternative to text-based and kernel-based methods.

%% file: results/main_result.tex
\begin{table*}[t]
\centering
\small
\renewcommand{\arraystretch}{1.2}
\setlength{\tabcolsep}{3pt}
\caption{%
Overall performance (\textbf{Hits@5}, \textbf{MRR}, and \textbf{NDCG}) on CC News and WikiText for text-, kernel-, and KGE-based methods under \textbf{Synonym}, \textbf{Context}, and \textbf{DIPPER} paraphrasing strengths. \textbf{Seen} means that the method directly accesses (or was trained with) the evaluation KG triples. For transductive KGE model, we use RotatE and for inductive KGE model, we use INGRAM. \textbf{Emb size} denotes the total parameter footprint used during similarity computation, calculated as \emph{(number of stored embedding vectors) $\times$ (embedding dimension)} across all original KGs in the corresponding dataset. Since VH Kernel and WL Kernel do not generate embedding vectors, we mark their \textbf{Emb size} with “N/A”. In the \textbf{Seen} column, $\bigcirc$ indicates that the method directly accesses or is trained on the evaluation KG triples, whereas $\times$ indicates that the evaluation KG triples are unseen during training. }
\label{tab:combined_results}
\vspace{-5mm}
\begin{tabular}{|l|l|c|r|cccccc|r|cccccc|}
\hline
\multirow{3}{*}{\textbf{Metric}}
  & \multirow{3}{*}{\textbf{Method}}
  & \multirow{3}{*}{\textbf{Seen}}
  & \multicolumn{7}{c|}{\textbf{CC News}}
  & \multicolumn{7}{c|}{\textbf{WikiText}}\\
\cline{4-17}
&&&
  \multicolumn{1}{c|}{\multirow{2}{*}{\shortstack{\textbf{Emb}\\\textbf{size}}}}
  & \multicolumn{2}{c}{\textbf{Synonym}}
  & \multicolumn{2}{c}{\textbf{Context}}
  & \multicolumn{2}{c|}{\textbf{DIPPER}}
  & \multicolumn{1}{c|}{\multirow{2}{*}{\shortstack{\textbf{Emb}\\\textbf{size}}}}
  & \multicolumn{2}{c}{\textbf{Synonym}}
  & \multicolumn{2}{c}{\textbf{Context}}
  & \multicolumn{2}{c|}{\textbf{DIPPER}}\\
\cline{5-10}\cline{12-17}
&&&&
  \textbf{30} & \textbf{60}
  & \textbf{30} & \textbf{60} & \textbf{60/0} & \textbf{60/20}
  &        & \textbf{30} & \textbf{60}
  & \textbf{30} & \textbf{60} & \textbf{60/0} & \textbf{60/20}\\
\hline
% ----------------------------------------------------------------
%                             Hits@5
% ----------------------------------------------------------------
\multirow{7}{*}{\textbf{Hits@5}}
& VH Kernel            & $\ocircle$
& N/A & 0.970 & 0.965 & 0.955 & 0.950 & 0.850 & 0.870
& N/A & 0.990 & 0.960 & 0.975 & 0.955 & 0.935 & 0.965\\
& WL Kernel            & $\ocircle$
& N/A & 0.920 & 0.930 & 0.930 & 0.900 & 0.805 & 0.815
& N/A & 0.980 & 0.965 & 0.960 & 0.940 & 0.905 & 0.945\\
\cline{2-17}
& SBERT                & $\times$
& 422,400 & 0.970 & 0.965 & 0.960 & 0.965 & \textbf{0.920} & \textbf{0.945}
& 495,360 & 0.955 & 0.955 & 0.985 & 0.965 & 0.940 & 0.940\\
\clineB{2-17}{3}
& EmbPairSim (RotatE)  & $\ocircle$
& 188,928 & \textbf{0.975} & \textbf{0.980} & \textbf{0.970} & \textbf{0.970} & 0.875 & 0.915
& 779,008 & \textbf{0.995} & \textbf{0.980} & \textbf{0.990} & \textbf{0.970} & \textbf{0.955} & \textbf{0.975}\\
& EmbPairSim (INGRAM)  & $\times$
& 188,928 & \textbf{0.975} & \textbf{0.980} & 0.965 & \textbf{0.970} & 0.875 & 0.915
& 779,008 & \textbf{0.995} & \textbf{0.980} & \textbf{0.990} & \textbf{0.970} & \textbf{0.955} & \textbf{0.975}\\
\cline{2-17}
& AvgEmbSim (RotatE)   & $\ocircle$
& 17,600 & 0.800 & 0.710 & 0.800 & 0.695 & 0.510 & 0.495
& 20,640 & 0.750 & 0.400 & 0.750 & 0.585 & 0.750 & 0.357\\
& AvgEmbSim (INGRAM)   & $\times$
& 17,600 & 0.790 & 0.705 & 0.745 & 0.585 & 0.410 & 0.445
& 20,640 & 0.675 & 0.565 & 0.755 & 0.670 & 0.400 & 0.365\\
\hline\hline
% ----------------------------------------------------------------
%                               MRR
% ----------------------------------------------------------------
\multirow{7}{*}{\textbf{MRR}}
& VH Kernel            & $\ocircle$
& N/A & \textbf{0.942} & 0.940 & 0.926 & 0.890 & 0.786 & 0.804
& N/A & 0.970 & 0.904 & 0.947 & 0.913 & 0.858 & 0.867\\
& WL Kernel            & $\ocircle$
& N/A & 0.881 & 0.876 & 0.885 & 0.821 & 0.717 & 0.729
& N/A & 0.960 & 0.914 & 0.931 & 0.902 & 0.812 & 0.869\\
\cline{2-17}
& SBERT                & $\times$
& 422,400 & 0.925 & 0.911 & 0.920 & 0.924 & \textbf{0.840} & \textbf{0.853}
& 495,360 & 0.935 & 0.909 & 0.963 & 0.924 & \textbf{0.909} & \textbf{0.901}\\
\clineB{2-17}{3}
& EmbPairSim (RotatE)  & $\ocircle$
& 188,928 & \textbf{0.942} & \textbf{0.953} & \textbf{0.928} & \textbf{0.928} & 0.830 & 0.834
& 779,008 & \textbf{0.988} & \textbf{0.952} & \textbf{0.975} & \textbf{0.945} & 0.887 & 0.897\\
& EmbPairSim (INGRAM)  & $\times$
& 188,928 & \textbf{0.942} & 0.946 & \textbf{0.928} & 0.889 & 0.827 & 0.833
& 779,008 & \textbf{0.988} & \textbf{0.945} & \textbf{0.973} & \textbf{0.945} & 0.879 & 0.889\\
\cline{2-17}
& AvgEmbSim (RotatE)   & $\ocircle$
& 17,600 & 0.751 & 0.655 & 0.739 & 0.625 & 0.438 & 0.437
& 20,640 & 0.698 & 0.314 & 0.748 & 0.543 & 0.492 & 0.314\\
& AvgEmbSim (INGRAM)   & $\times$
& 17,600 & 0.731 & 0.650 & 0.703 & 0.526 & 0.362 & 0.383
& 20,640 & 0.620 & 0.461 & 0.667 & 0.601 & 0.345 & 0.300\\
\hline\hline
% ----------------------------------------------------------------
%                               NDCG
% ----------------------------------------------------------------
\multirow{7}{*}{\textbf{NDCG}}
& VH Kernel            & $\ocircle$
& N/A & 0.953 & 0.952 & 0.940 & 0.912 & 0.824 & 0.835
& N/A & 0.978 & 0.923 & 0.960 & 0.930 & 0.886 & 0.896\\
& WL Kernel            & $\ocircle$
& N/A & 0.905 & 0.903 & 0.908 & 0.858 & 0.771 & 0.777
& N/A & 0.969 & 0.930 & 0.946 & 0.921 & 0.850 & 0.895\\
\cline{2-17}
& SBERT                & $\times$
& 422,400 & 0.943 & 0.931 & 0.938 & 0.942 & \textbf{0.876} & \textbf{0.887}
& 495,360 & 0.949 & 0.928 & 0.971 & 0.941 & \textbf{0.928} & \textbf{0.923}\\
\clineB{2-17}{3}
& EmbPairSim (RotatE)  & $\ocircle$
& 188,928 & \textbf{0.954} & \textbf{0.963} & \textbf{0.943} & \textbf{0.949} & 0.861 & 0.861
& 779,008 & \textbf{0.991} & \textbf{0.960} & \textbf{0.981} & \textbf{0.954} & 0.907 & 0.918\\
& EmbPairSim (INGRAM)  & $\times$
& 188,928 & \textbf{0.954} & 0.957 & \textbf{0.943} & 0.913 & 0.857 & 0.859
& 779,008 & \textbf{0.991} & 0.954 & \textbf{0.980} & \textbf{0.954} & 0.900 & 0.912\\
\cline{2-17}
& AvgEmbSim (RotatE)   & $\ocircle$
& 17,600 & 0.794 & 0.708 & 0.784 & 0.686 & 0.519 & 0.518
& 20,640 & 0.748 & 0.410 & 0.835 & 0.606 & 0.566 & 0.398\\
& AvgEmbSim (INGRAM)   & $\times$
& 17,600 & 0.782 & 0.715 & 0.759 & 0.611 & 0.469 & 0.483
& 20,640 & 0.689 & 0.559 & 0.729 & 0.672 & 0.451 & 0.411\\
\hline
\end{tabular}
\vspace{-2mm}
\end{table*}

%% file: analysis/kge_model/figure.tex
\begin{figure*}[h]
    \centering
    \includegraphics[width=\textwidth]{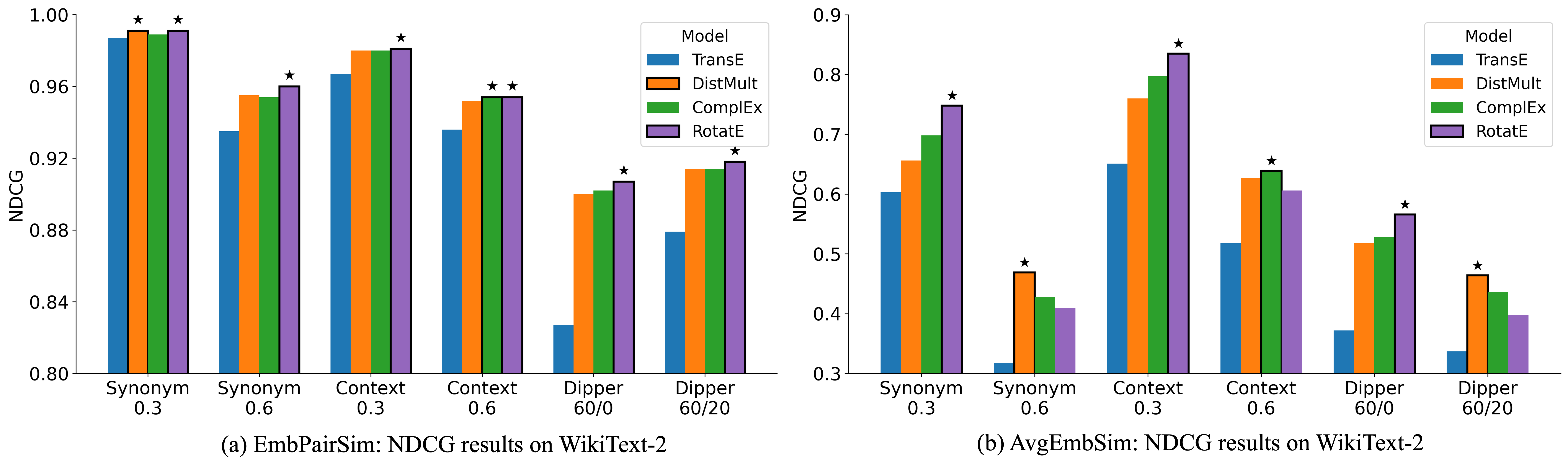}
    \vspace{-5mm}
    \small\caption{NDCG scores of \textit{EmbPairSim} and \textit{AvgEmbSim} on \textit{WikiText} using different transductive KGE models across all modification settings.
    For each setting, the highest NDCG score is marked with a black outline and a star.}
    \vspace{-5pt}
    \label{fig:models}
\end{figure*}

%% file: results/ablation_study/AvgEmbSim.tex
\begin{table}[t]

\centering
% \scriptsize 
\small
\caption{
Ablation study on \textit{AvgEmbSim} using the RotatE model on WikiText-2, comparing setting with (\textbf{w/ freq}) and without (\textbf{w/o freq}) entity frequency information.
}

\label{tab:ablation_avg}
\resizebox{\linewidth}{!}{
\begin{tabular}{|ll|cccccc|}
\hline
\textbf{Metric} & \textbf{Option} 
& \multicolumn{2}{c}{\textbf{Synonym}} 
& \multicolumn{2}{c}{\textbf{Context}} 
& \multicolumn{2}{c|}{\textbf{DIPPER}} \\
\cline{3-8}
& & \textbf{30} & \textbf{60} & \textbf{30} & \textbf{60} & \textbf{60/0} & \textbf{60/20} \\
\hline

% ======================== Hits@5 ========================
\multirow{2}{*}{Hits@5}
& w/ freq
& \textbf{0.750} & 0.400 & \textbf{0.750} & \textbf{0.585} & \textbf{0.750} & \textbf{0.357} \\
& w/o freq
& 0.675 & \textbf{0.595} & 0.655 & 0.560 & 0.385 & 0.320 \\

% ======================== MRR ========================
\hline
\multirow{2}{*}{MRR}
& w/ freq
& \textbf{0.698} & 0.314 & \textbf{0.748} & \textbf{0.543} & \textbf{0.492} & \textbf{0.314} \\
& w/o freq
& 0.600 & \textbf{0.524} & 0.609 & 0.470 & 0.329 & 0.268 \\

% ======================== NDCG ========================
\hline
\multirow{2}{*}{NDCG}
& w/ freq
& \textbf{0.748} & 0.410 & \textbf{0.835} & \textbf{0.606} & \textbf{0.566} & \textbf{0.398} \\
& w/o freq 
& 0.666 & \textbf{0.597} & 0.672 & 0.556 & 0.429 & 0.378 \\
\hline
\end{tabular}}

\end{table}

%% file: results/ablation_study/EmbPairSim-INGRAM.tex
\begin{table}[t]
\vspace{-7pt}
\centering
% \scriptsize 
\small
\caption{
Ablation study on \textit{EmbPairSim} using the INGRAM model on WikiText-2, comparing setting with (\textbf{w/ cent}) and without (\textbf{w/o cent}) mean centering.
}
\label{tab:ablation_pair_ingram}
\resizebox{\linewidth}{!}{
\begin{tabular}{|ll|cccccc|}
\hline
\textbf{Metric} & \textbf{Option} 
& \multicolumn{2}{c}{\textbf{Synonym}} 
& \multicolumn{2}{c}{\textbf{Context}}
& \multicolumn{2}{c|}{\textbf{DIPPER}} \\
\cline{3-8}
& & \textbf{30} & \textbf{60} & \textbf{30} & \textbf{60} & \textbf{60/0} & \textbf{60/20} \\
\hline

% ======================== Hits@5 ========================
\multirow{2}{*}{{Hits@5}}
& w/ cent
% & \textbf{0.970} & \textbf{0.905} & \textbf{0.970} & \textbf{0.900} & \textbf{0.855} & \textbf{0.855}  \\
& \textbf{0.995} & \textbf{0.980} & \textbf{0.990} & \textbf{0.970} & \textbf{0.955} & \textbf{0.975} \\

& w/o cent
% & 0.000 & 0.000 & 0.000 & 0.000 & 0.000 & 0.000 \\
& 0.000 & 0.000 & 0.000 & 0.000 & 0.000 & 0.000 \\

% ======================== MRR ========================
\hline
\multirow{2}{*}{{MRR}}
& w/ cent
% & \textbf{0.947} & \textbf{0.854} & \textbf{0.928} & \textbf{0.810} & \textbf{0.806} & \textbf{0.788}  \\
& \textbf{0.988} & \textbf{0.945} & \textbf{0.973} & \textbf{0.945} & \textbf{0.879} & \textbf{0.889} \\

& w/o cent
% & 0.000 & 0.000 & 0.001 & 0.000 & 0.000 & 0.000 \\
& 0.002 & 0.002 & 0.002 & 0.002 & 0.002 & 0.002 \\

% ======================== NDCG ========================
\hline
\multirow{2}{*}{{NDCG}}
& w/ cent
% & \textbf{0.957} & \textbf{0.875} & \textbf{0.943} & \textbf{0.845} & \textbf{0.830} & \textbf{0.821}  \\
& \textbf{0.991} & \textbf{0.954} & \textbf{0.980} & \textbf{0.954} & \textbf{0.900} & \textbf{0.912} \\

& w/o cent
% & 0.088 & 0.088 & 0.089 & 0.087 & 0.087 & 0.088 \\
& 0.108 & 0.108 & 0.109 & 0.108 & 0.108 & 0.109 \\
\hline
\end{tabular}}
\vspace{-5pt}

\end{table}

%% file: results/relation/EmbPairSim.tex
\begin{table}[t]
% \scriptsize 
\small
\centering
\caption{Comparison of \textit{EmbPairSim} (\textit{pair}) and \textit{AvgEmbSim} (\textit{avg}) on WikiText-2 (RotatE embeddings).
\textbf{w/o rel}: entity-only; \textbf{w/ rel}: entity + relation embeddings.}
\label{tab:relation_pair_avg}

\setlength{\tabcolsep}{3pt}%
\resizebox{\linewidth}{!}{
\begin{tabular}{|llc|cccccc|}
\hline
\textbf{Metric} & \textbf{Option} & \textbf{Method} &
\multicolumn{2}{c}{\textbf{Synonym}} & \multicolumn{2}{c}{\textbf{Context}} & \multicolumn{2}{c|}{\textbf{Dipper}} \\
\cline{4-9}
& & & \textbf{30} & \textbf{60} & \textbf{30} & \textbf{60} & \textbf{60/0} & \textbf{60/20} \\
\hline
\multirow{2}{*}{Hits@5}
& w/o rel & pair & \textbf{0.995} & \textbf{0.980} & \textbf{0.990} & \textbf{0.970} & \textbf{0.955} & \textbf{0.975} \\
& w/ rel  & pair & \textbf{0.995} & 0.955 & \textbf{0.990} & 0.955 & 0.910 & 0.925 \\
\hline
\multirow{2}{*}{MRR}
& w/o rel & pair & \textbf{0.988} & \textbf{0.952} & \textbf{0.975} & \textbf{0.945} & \textbf{0.887} & \textbf{0.897} \\
& w/ rel  & pair & \textbf{0.988} & 0.921 & 0.973 & 0.905 & 0.811 & 0.816 \\
\hline
\multirow{2}{*}{NDCG}
& w/o rel & pair & \textbf{0.991} & \textbf{0.960} & \textbf{0.981} & \textbf{0.954} & \textbf{0.907} & \textbf{0.918} \\
& w/ rel  & pair & \textbf{0.991} & 0.936 & 0.980 & 0.924 & 0.850 & 0.856 \\
\hline
\multirow{2}{*}{Hits@5}
& w/o rel & avg & \textbf{0.750} & \textbf{0.400} & \textbf{0.750} & \textbf{0.585} & \textbf{0.750} & \textbf{0.357} \\
& w/ rel  & avg & 0.250 & 0.180 & 0.330 & 0.190 & 0.145 & 0.095 \\
\hline
\multirow{2}{*}{MRR}
& w/o rel & avg & \textbf{0.698} & \textbf{0.314} & \textbf{0.748} & \textbf{0.543} & \textbf{0.492} & \textbf{0.314} \\
& w/ rel  & avg & 0.229 & 0.170 & 0.282 & 0.180 & 0.122 & 0.096 \\
\hline
\multirow{2}{*}{NDCG}
& w/o rel & avg & \textbf{0.748} & \textbf{0.410} & \textbf{0.835} & \textbf{0.606} & \textbf{0.566} & \textbf{0.398} \\
& w/ rel  & avg & 0.345 & 0.290 & 0.392 & 0.298 & 0.247 & 0.219 \\
\hline
\end{tabular}}
\vspace{-2mm}
\end{table}

%% file: results/LM/EmbPairSim/trained.tex
\begin{table}[t]
\centering
\small
% \scriptsize                
\caption{EmbPairSim performance on WikiText-2 using Word2Vec and FastText embeddings. In the \textbf{Pretrained} column, $\bigcirc$ indicates that externally pretrained embeddings are used, whereas $\times$ indicates that the embeddings are trained only on the WikiText-2 corpus. }
\label{tab:we_pair}
\setlength{\tabcolsep}{3pt}
\resizebox{\linewidth}{!}{
\begin{tabular}{|llc|cccccc|}
\hline
\textbf{Metric} & \textbf{Model} & \textbf{Pretrained} & \multicolumn{2}{c}{Synonym} & \multicolumn{2}{c}{Context} & \multicolumn{2}{c|}{Dipper} \\
\cline{4-9}
& & & \textbf{30} & \textbf{60} & \textbf{30} & \textbf{60} & \textbf{60/0} & \textbf{60/20} \\
\hline

% Hits@5 (non-pretrained)
\multirow{3}{*}{Hits@5}
& FastText  & $\times$ & 0.990 & 0.940 & 0.985 & 0.955 & 0.915 & 0.955 \\
& Word2Vec  & $\times$ & 0.680 & 0.445 & 0.635 & 0.495 & 0.300 & 0.285 \\
& RotatE    & $\times$ & \textbf{0.995} & \textbf{0.980} & \textbf{0.990} & \textbf{0.970} & \textbf{0.955} & \textbf{0.975} \\

\hline
% MRR (non-pretrained)
\multirow{3}{*}{MRR}
& FastText  & $\times$ & 0.972 & 0.865 & 0.957 & 0.903 & 0.778 & 0.793 \\
& Word2Vec  & $\times$ & 0.399 & 0.258 & 0.375 & 0.344 & 0.170 & 0.186 \\
& RotatE    & $\times$ & \textbf{0.988} & \textbf{0.952} & \textbf{0.975} & \textbf{0.945} & \textbf{0.887} & \textbf{0.897} \\

\hline
% NDCG (non-pretrained)
\multirow{3}{*}{NDCG}
& FastText  & $\times$ & 0.979 & 0.893 & 0.968 & 0.922 & 0.823 & 0.839 \\
& Word2Vec  & $\times$ & 0.531 & 0.399 & 0.510 & 0.477 & 0.321 & 0.334 \\
& RotatE    & $\times$ & \textbf{0.991} & \textbf{0.960} & \textbf{0.981} & \textbf{0.954} & \textbf{0.907} & \textbf{0.918} \\

\hline
% Hits@5 (pretrained)
\multirow{3}{*}{Hits@5}
& FastText  & $\ocircle$ & \textbf{0.995} & 0.975 & 0.975 & 0.965 & 0.950 & 0.965 \\
& Word2Vec  & $\ocircle$ & 0.755 & 0.530 & 0.695 & 0.570 & 0.575 & 0.590 \\
& INGRAM    & $\ocircle$ & \textbf{0.995} & \textbf{0.980} & \textbf{0.990} & \textbf{0.970} & \textbf{0.955} & \textbf{0.975} \\

\hline
% MRR (pretrained)
\multirow{3}{*}{MRR}
& FastText  & $\ocircle$ & 0.974 & 0.925 & 0.959 & 0.921 & 0.847 & 0.870 \\
& Word2Vec  & $\ocircle$ & 0.592 & 0.381 & 0.523 & 0.450 & 0.434 & 0.413 \\
& INGRAM    & $\ocircle$ & \textbf{0.988} & \textbf{0.945} & \textbf{0.973} & \textbf{0.945} & \textbf{0.879} & \textbf{0.889} \\

\hline
% NDCG (pretrained)
\multirow{3}{*}{NDCG}
& FastText  & $\ocircle$ & 0.980 & 0.940 & 0.968 & 0.936 & 0.876 & 0.897 \\
& Word2Vec  & $\ocircle$ & 0.680 & 0.500 & 0.626 & 0.557 & 0.541 & 0.528 \\
& INGRAM    & $\ocircle$ & \textbf{0.991} & \textbf{0.954} & \textbf{0.980} & \textbf{0.954} & \textbf{0.900} & \textbf{0.912} \\
\hline
\end{tabular}}
\vspace{-3mm}
\end{table}

%% file: conclusion.tex
\section{Conclusion}\label{sec:conclusion}
We presented a systematic approach to compute semantic similarity between KGs using SBERT, graph kernels, and KGE-based methods. To leverage KGE more effectively, we proposed two complementary functions, \textit{EmbPairSim} and \textit{AvgEmbSim}. Experimental results show that KGE can capture semantic information, supporting the view that a KG can be interpreted hierarchically like a document, with entities and relations as words and triples and subgraphs forming higher-level meaning. Future work should explore end-to-end graph neural architectures that better encode structural patterns and relational dependencies for richer graph-level semantics.
% In this study, we systematically compute semantic similarity between knowledge graphs using Sentence BERT, graph kernels, and KGE models, and evaluate their performance through a semantic matching task. To lift entity embeddings from KGE models, we develop two complementary KGE-based functions: EmbPairSim, and AvgEmbSim. The results from our experiments show the effectiveness of KGE, indicating that a knowledge graph can be hierarchically interpreted like a document—where entities and relations function as words and phrases, and triples and subgraphs correspond to sentences and paragraphs, collectively forming graph-level meaning.

% \section*{Acknowledgements}
% This work was supported by the IITP, NRF, and KOCCA grants funded by the Korean government (RS-2019-II190421, IITP-2025-RS-2020-II201821, RS-2024-00438686, RS-2024-00436936, RS-2023-00225441, RS-2024-00448809, IITP-2025-RS-2024-00360227, RS-2025-02218768, RS-2024-00333068).

% \section{Limitation and Future Work}
% % Limitation & Future work 
% While our results demonstrate the feasibility of using KGE for semantic similarity, the limited performance \textit{AvgEmbSim} and the performance drop of \textit{EmbPairSim} on Dipper Paraphraser cases highlight that simple statistical aggregation struggles to capture the full graph context. Future work should explore end-to-end graph neural architectures that better encode structural patterns and relational dependencies for richer graph-level semantics.

%% file: appendix.tex
\section{Examples of KG Information at Different Granularities} 
\label{sec:multi_level_kg_examples}
In scholarly KGs like ORKG~\cite{ScholarlyKG}, contribution-centered subgraphs represent contributions and are used to compare related literature, and the entire KG represents a scholarly knowledge base and is used to support research comparison and thematic review across papers or research areas, while entities and relations represent scholarly units and their connections and triples encode structured scholarly statements.
In event-centric KGs like EventKG~\cite{EventKG}, event-centered subgraphs represent event contexts and are used to support event exploration and timeline generation, and the entire KG represents a temporal event knowledge base and is used for event-centric analysis, while entities and relations represent events, participants, locations, times, and temporal or semantic connections and triples encode individual event facts.
In scene graphs~\cite{VisualGenome}, region-level subgraphs represent image regions and are used for region-level visual understanding, and the entire KG represents image-level visual content and is used for image-level visual understanding, while entities and relations represent visual objects, attributes, and interactions and triples encode local visual relationships.
In encyclopedic KGs like Wikidata~\cite{wikidata}, item- or topic-centered subgraphs represent item-level knowledge and are used to retrieve item information, statements, and provenance, and the entire KG represents an encyclopedic knowledge base and is used for structured search over entities and attributes across domains, while entities and relations represent encyclopedic units and their properties and triples correspond to item-property-value statements.
In biomedical KGs like BioKG~\cite{BioKG}, disease-, drug-, or pathway-centered subgraphs represent biomedical contexts and are used to trace relations among diseases, drugs, genes, and pathways for biomedical knowledge discovery, and the entire KG represents a biomedical knowledge base and is used for causal inference, drug repurposing, and drug target identification, while entities and relations represent biomedical units and their associations and triples encode individual biomedical facts.
In legal KGs like LegalKG~\cite{LegalKG}, case- or regulation-centered subgraphs represent legal contexts and are used to trace dependencies between legal provisions and judicial decisions, and the entire KG represents a legal knowledge base and is used to search, connect, and process legal norms and court decisions, while entities and relations represent legal units and their dependencies and triples encode individual legal statements.